\documentclass[10pt, a4paper]{article}
\usepackage{lrec}
\usepackage[latin1]{inputenc}
\usepackage{graphicx}
\usepackage{tabularx}
\usepackage{soul}

\usepackage{microtype}
\usepackage{url}
\usepackage{tabularx}
\usepackage{multirow}
\usepackage{xcolor}
\usepackage{scalefnt}
\usepackage{booktabs}
\newcommand{\rt}[1]{\rotatebox{90}{#1}}

\usepackage{pifont}
\usepackage{xspace}
\newcommand{\eg}{\textit{e.\,g.}\xspace}

\newcommand{\F}{$\textrm{F}_1$\xspace}

\usepackage{multirow}
\newcommand{\mycross}{\textcolor{black!50!white}{\ding{55}}}
\newcommand{\mycheck}{\ding{51}}
\newcommand{\ccol}[1]{\multicolumn{1}{c}{#1}}

\title{GoodNewsEveryone: A Corpus of News Headlines Annotated with \\ Emotions, Semantic Roles, and Reader Perception}
\name{Laura Bostan, Evgeny Kim, Roman Klinger}

\address{%
  Institut f{\"u}r Maschinelle Sprachverarbeitung, %
  Universit{\"a}t Stuttgart\\
  Pfaffenwaldring 5b, 70569 Stuttgart, Germany \\
  \{laura.bostan, evgeny.kim, roman.klinger\}@ims.uni-stuttgart.de\\
}

\abstract{Most research on emotion analysis from text focuses on the
  task of emotion classification or emotion intensity
  regression. Fewer works address emotions as a phenomenon to be
  tackled with structured learning, which can be explained by the lack
  of relevant datasets. We fill this gap by releasing a dataset of
  5000 English news headlines annotated via crowdsourcing with their
  associated emotions, the corresponding emotion experiencers and
  textual cues, related emotion causes and targets, as well as the
  reader's perception of the emotion of the headline. This annotation
  task is comparably challenging, given the large number of classes
  and roles to be identified. We therefore propose a multiphase
  annotation procedure in which we first find relevant instances with
  emotional content and then annotate the more fine-grained
  aspects. Finally, we develop a baseline for the task of automatic
  prediction of semantic role structures and discuss the results. The
  corpus we release enables further research on emotion
  classification, emotion intensity prediction, emotion cause
  detection, and supports further qualitative studies.
}

\begin{document}

\maketitleabstract

\section{Introduction}

Research in emotion analysis from text focuses on mapping words, sentences, or
documents to emotion categories based on the models of \newcite{Ekman1992} or
\newcite{Plutchik2001}, which propose the emotion classes of \textit{joy,
sadness, anger, fear, trust, disgust, anticipation} and \textit{surprise}.
Emotion analysis has been applied to a variety of tasks including large scale
social media mining \cite{Xuan2013}, literature analysis
\cite{Reagan2016,Kim2019}, lyrics and music analysis
\cite{Mihalcea2012,dodds2010measuring}, and the analysis of the development of
emotions over time \cite{Hellrich2019}.

There are at least two types of questions that cannot yet be answered by these
emotion analysis systems. Firstly, such systems do not often explicitly model
the perspective of understanding the written discourse (reader, writer, or the
text's point of view). For example, the headline ``Djokovic happy to carry on
cruising'' \cite{headline2} contains an explicit mention of \textit{joy}
carried by the word ``happy''. However, it may evoke different emotions in a
reader (\eg, when the reader is a supporter of Roger Federer), and the same
applies to the author of the headline. To the best of our knowledge, only one
work considers this point \cite{Buechel2017b}. Secondly, the structure that can
be associated with the emotion description in text is not uncovered. Questions
like ``Who feels a particular emotion?'' or ``What causes that emotion?''
remain unaddressed. There has been almost no work in this direction, with only
a few exceptions in English \cite{Kim2018,Mohammad2014} and Mandarin
\cite{Xu2019,Ding2019}.

With this work, we argue that emotion analysis would benefit from a more
fine-grained analysis that considers the full structure of an emotion, similar
to the research in aspect-based sentiment analysis
\cite{Wang2016,Ma2018,Xue2018,Sun2019}.
\begin{figure*}
  \centering
  \includegraphics[width=0.95\linewidth]{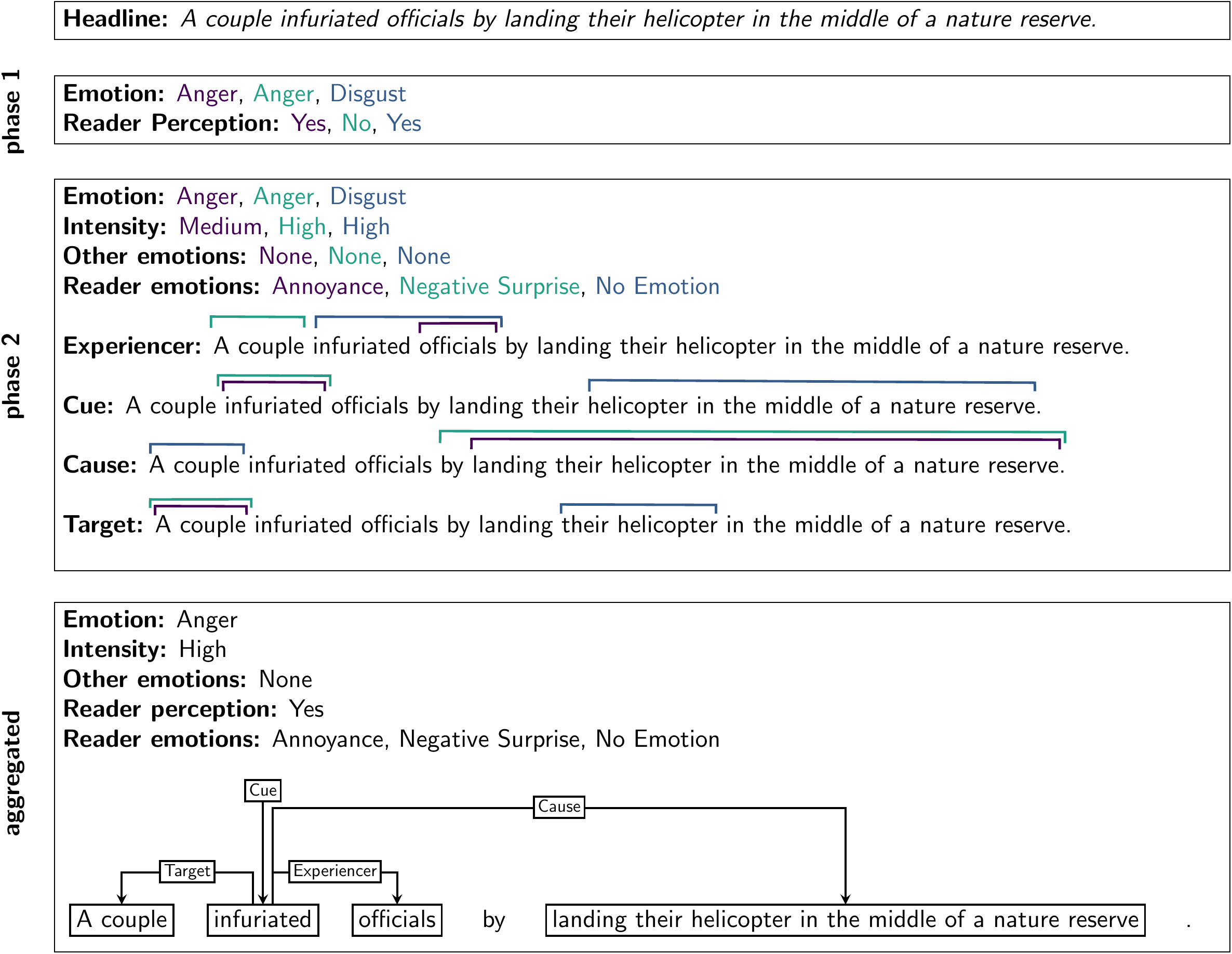}
  \caption{Example of an annotated headline from our dataset. Each color represents an annotator.}
  \label{example-details}
\end{figure*}
Consider the headline: ``A couple infuriated officials by landing their
helicopter in the middle of a nature reserve'' \cite{headline1} depicted in
Figure~\ref{example-details}. One could mark ``officials'' as the experiencer,
``a couple'' as the target, and ``landing their helicopter in the middle of a
nature reserve'' as the cause of \textit{anger}. Now let us imagine that the
headline starts with ``A \textit{cheerful} couple'' instead of ``A couple''. A
simple approach to emotion detection based on cue words will capture that this
sentence contains descriptions of \textit{anger} (``infuriated'') and
\textit{joy} (``cheerful''). It would, however, fail in attributing correct
roles to the couple and the officials. Thus, the distinction between their
emotional experiences would remain hidden from us.

In this study, we focus on an annotation task to develop a dataset
that would enable addressing the issues raised above. Specifically, we
introduce the corpus \textit{GoodNewsEveryone}, a novel dataset of
English news headlines collected from 82 different sources most of
which are analyzed in the Media Bias Chart \cite{Otero2018} annotated
for emotion class, emotion intensity, semantic roles (experiencer,
cause, target, cue), and reader perspective. We use semantic roles,
since identifying who feels what and why is essentially a semantic
role labeling task \cite{Gildea2002}. The roles we consider are a
subset of those defined for the semantic frame for ``Emotion'' in
FrameNet \cite{Baker1998}.

We focus on news headlines due to their brevity and density of contained
information. Headlines often appeal to a reader's emotions and hence are a
potentially good source for emotion analysis. Besides, news headlines are
easy-to-obtain data across many languages, void of data privacy issues
associated with social media and microblogging.

Further, we opt for a crowdsourcing setting in contrast to an
expert-based setting to obtain data annotated that is to a lesser
extend influenced by individual opinions of a low number of
annotators. Besides, our previous work showed that it is comparably
hard to reach an acceptable agreement in such tasks even under close
supervision \cite{Kim2018}.

To summarize, our main contributions in this paper are, (1), that we
present the first resource of news headlines annotated for emotions,
cues, intensities, experiencers, causes, targets, and reader emotion,
(2), design a two-phase annotation procedure for emotion structures
via crowdsourcing, and, (3), provide results of a baseline model to
predict such roles in a sequence labeling setting. We provide our
annotation guidelines and annotations at
\url{http://www.ims.uni-stuttgart.de/data/goodnewseveryone}.

\section{Related Work} 
Our annotation and modelling project is inspired by emotion
classification and intensity prediction as well as role labeling and
resources which were prepared for these tasks. We therefore look into
each of these subtasks and explain how they are related to our new
corpus.

\subsection{Emotion Classification} Emotion classification deals with mapping
words, sentences, or documents to a set of emotions following psychological
models such as those proposed by \newcite{Ekman1992} (\textit{anger,
  disgust, fear, joy, sadness}, and \textit{surprise}) or \newcite{Plutchik2001}; or
continuous values of \textit{valence}, \textit{arousal} and \textit{dominance}
\cite{Russell1980}.

Datasets for those tasks can be created in different ways.  One way to
create annotated datasets is via \textit{expert annotation}
\cite{Aman2007,Strapparava2007,Ghazi2015,Schuff2017,Buechel2017b}. A
special case of this procedure has been proposed by the creators of
the ISEAR dataset who make use of self-reporting instead, where
subjects are asked to describe situations associated with a specific
emotion \cite{Scherer1997}.

\emph{Crowdsourcing} is another popular way to acquire human judgments
\cite{Mohammad2012,Mohammad2014,Mohammad2014,Abdul2017,Mohammad2018},
for instance on Amazon Mechanical Turk or Figure Eight (previously
known as Crowdflower). \newcite{Troiano2019} recently published a data
set which combines the idea of requesting self-reports (by experts in
a lab setting) with the idea of using crowdsourcing. They extend their
data to German reports (next to English) and validate each instance,
again, via crowdsourcing.

Lastly, social network platforms play a central role in data
acquisition with distant supervision, because they provide a cheap way
to obtain large amounts of noisy data
\cite{Mohammad2012,Mohammad2014,Mohammad2015b,Liu2017b}.

We show an overview of available resources in
Table~\ref{tab:resources}. Further, more details on previous work can
for instance be found in \newcite{Bostan2018}.

\begin{table*}
\centering\small
\begin{tabular}{ll@{\hskip 0.1in}l@{\hskip 0.3in}rrrrr@{\hskip 0.1in}r@{\hskip 0.1in}l}
\toprule
& Dataset &  Emotion Annotation &
\rt{Int.} &
\rt{Cue} &
\rt{Exp.} &
\rt{Cause} &
\rt{Target} &
\ccol{Size} &  Source \\
\cmidrule(rl){2-2}\cmidrule(r){3-3}\cmidrule(r){4-4}\cmidrule(lr){5-5}\cmidrule(lr){6-6}\cmidrule(lr){7-7}\cmidrule(lr){8-8}\cmidrule(r){9-9}\cmidrule{10-10}
\multirow{13}{*}{\rt{Emotion \& Intensity Classification}}
& ISEAR & Ekman + \{shame, guilt\}  &\mycross
&\mycross&\mycross&\mycross &\mycross & 7,665 & Scherer et al. \shortcite{Scherer1997} \\
& Tales & Ekman & \mycross &\mycross&\mycross& \mycross &\mycross & 15,302  & \newcite{Ovesdotter2005} \\
& AffectiveText & Ekman + \{valence\} &\mycross &\mycross&\mycross&
\mycross &\mycross &  1,250 & Strapparava et al. \shortcite{Strapparava2007} \\
& TEC & Ekman + \{$\pm$surprise\} &\mycross &\mycross&\mycross& \mycross &\mycross  & 21,051 &  \newcite{Mohammad2015a} \\
& fb-valence-arousal  & VA  &\mycross &\mycross&\mycross& \mycross &\mycross& 2,895 &  \newcite{Preotiuc2016}  \\
& EmoBank & VAD   &\mycross &\mycross&\mycross& \mycross &\mycross&  10,548  &  \newcite{Buechel2017a} \\
& DailyDialogs  & Ekman &\mycross &\mycross&\mycross& \mycross &\mycross &  13,118 &  \newcite{Li2017}  \\
& Grounded-Emotions & Joy \& Sadness   &\mycross &\mycross&\mycross& \mycross &\mycross & 2,585 & \newcite{Liu2017b} \\
& SSEC & Plutchik &\mycross &\mycross&\mycross& \mycross &\mycross &  4,868 & \newcite{Schuff2017} \\
& EmoInt & Ekman $-$ \{disgust, surprise\}  & \mycheck &\mycross&\mycross& \mycross &\mycross& 7,097  & Mohammad et al. \shortcite{Mohammad2017} \\
& Multigenre & Plutchik  & \mycross &\mycross&\mycross& \mycross &\mycross& 17,321  & \newcite{Tafreshi2018} \\
& The Affect in Tweets  & Others & \mycheck &\mycross&\mycross&\mycross &\mycross &  11,288 & Mohammad \shortcite{TweetEmo}\\ 
& EmoContext & Joy, Sadness, Anger \& Others & \mycross &\mycross&\mycross& \mycross &\mycross & 30,159  &  \newcite{Chatterjee2019}\\
& MELD & Ekman + Neutral  &\mycross &\mycross&\mycross& \mycross &\mycross  &  13,000 & \newcite{Poria2019} \\
& enISEAR & Ekman + \{shame, guilt\} &\mycross &\mycross&\mycross& \mycross &\mycross  & 1,001  & \newcite{Troiano2019} \\
\cmidrule(rl){1-2}\cmidrule(r){3-3}\cmidrule(r){4-4}\cmidrule(lr){5-5}\cmidrule(lr){6-6}\cmidrule(lr){7-7}\cmidrule(lr){8-8}\cmidrule(r){9-9}\cmidrule{10-10}
\multirow{7}{*}{\rt{Roles}}
& Blogs  & Ekman + \{mixed, noemo\}  &\mycheck &\mycheck&\mycross&\mycross &\mycross &  5,025 & Aman et al. \shortcite{Aman2007} \\
& Emotion-Stimulus  & Ekman + \{shame\} & \mycross &\mycross& \mycross & \mycheck & \mycross &  2,414  & \newcite{Ghazi2015} \\
& EmoCues & 28 emo categories & \mycross & \mycheck &\mycross&\mycross& \mycross & 15,553 & \newcite{Yan2016emocues}  \\
& Electoral-Tweets  & Plutchik & \mycross &\mycheck &\mycheck & \mycheck & \mycheck &   4,058 &  \newcite{Mohammad2014} \\
& REMAN & Plutchik + \{other\}  & \mycross & \mycheck & \mycheck & \mycheck  & \mycheck & 1,720 &  \newcite{Kim2018} \\
 & \textbf{GoodNewsEveryone} & \textbf{extended Plutchik}  & \mycheck & \mycheck & \mycheck & \mycheck & \mycheck & \textbf{5,000} & \textbf{Bostan et. al (2020)} \\
\bottomrule
\end{tabular}
\caption{Related resources for emotion analysis in English.}
\label{tab:resources}
\end{table*}

\subsection{Emotion Intensity}

In emotion intensity prediction, the term \textit{intensity} refers to
the \textit{degree} an emotion is experienced. For this task, there
are only a few datasets available. To our knowledge, the first dataset
annotated for emotion intensity is by \newcite{Aman2007}, who ask
experts to map textual spans to a set of predefined categories of
emotion intensity (\textit{high, moderate}, and
\textit{low}). Recently, new datasets were released for the EmoInt
shared tasks \cite{Mohammad2017,Mohammad2018}, both annotated via
crowdsourcing through best-worst scaling.

\subsection{Cue or Trigger Words}

The task of finding a function that segments a textual input and finds
the span indicating an emotion category is less researched. First work
that annotated cues was done manually by one expert and three
annotators on the domain of blog posts \cite{Aman2007}.
\newcite{Mohammad2014} annotate the cues of emotions in a corpus of
4,058 electoral tweets from the US via crowdsourcing. Similar in
annotation procedure, \newcite{Yan2016emocues} curate a corpus of
15,553 tweets and annotate it with 28 emotion categories, valence,
arousal, and cues.

To the best of our knowledge, there is only one work \cite{Kim2018}
that leverages the annotations for cues and considers the task of
emotion detection where the exact spans that represent the cues need
to be predicted.

\subsection{Emotion Cause Detection}

Detecting the cause of an expressed emotion in text received
relatively little attention, compared to emotion detection. There are
only few works on English that focus on creating resources to tackle
this task \cite{Ghazi2015,Mohammad2014,Kim2018,Gao2015}. The task can
be formulated in different ways. One is to define a closed set of
potential causes after annotation. Then, cause detection is a
classification task \cite{Mohammad2014}.  Another setting is to find
the cause in the text without sticking to clause boundaries. This is
formulated as segmentation or clause classification on the token level
\cite{Ghazi2015,Kim2018}. Finding the cause of an emotion is widely
researched on Mandarin in both resource creation and methods. Early
works build on rule-based systems \cite{Lee2010,Lee2010text,Chen2010},
which examine correlations between emotions and cause events in terms
of linguistic cues. The works that follow up focus on both methods and
corpus construction, showing large improvements over the early works
\cite{Li2014text,Gui2014,Gao2015,Gui2016,Gui2017,Xu2017,Cheng2017,Chen2018,Ding2019}.
The most recent work on cause extraction is being done on Mandarin and
formulates the task jointly with emotion detection
\cite{Xu2019,Xia2019a,Xia2019b}. With the exception of
\newcite{Mohammad2014} who are annotating via crowdsourcing, all other
datasets are manually labeled by experts, usually using the W3C
Emotion Markup
Language\footnote{\url{https://www.w3.org/TR/emotionml/}, last
  accessed Nov 27 2019}.

\subsection{Semantic Role Labeling of Emotions}

Semantic role labeling in the context of emotion analysis deals with extracting
who feels (\textit{experiencer}) which emotion (\textit{cue}, \textit{class}),
towards whom the emotion is directed (\textit{target}), and what is the event
that caused the emotion (\textit{stimulus}). The relations are defined akin to
FrameNet's Emotion frame \cite{Baker1998}.

There are two works that work on annotation of semantic roles in the
context of emotion. Firstly, \newcite{Mohammad2014} annotate a dataset
of 4,058 tweets via crowdsourcing. The tweets were published before
the U.S. presidential elections in 2012. The semantic roles considered
are the experiencer, the stimulus, and the target. However, in the
case of tweets, the experiencer is mostly the author of the
tweet. Secondly, \newcite{Kim2018} annotate and release REMAN
(Relational EMotion ANnotation), a corpus of 1,720 paragraphs based on
Project Gutenberg. REMAN was manually annotated for spans which
correspond to emotion cues and entities/events in the roles of
experiencers, targets, and causes of the emotion. They also provide
baseline results for the automatic prediction of these structures and
show that their models benefit from joint modeling of emotions with
its roles in all subtasks. Our work follows in motivation
\newcite{Kim2018} and in procedure \newcite{Mohammad2014}.

\subsection{Reader vs.  Writer vs. Text Perspective}

Studying the impact of different annotation perspectives is another
little explored area. There are few exceptions in sentiment analysis
which investigate the relation between sentiment of a blog post and
the sentiment of their comments \cite{Tang2012} or model the emotion
of a news reader jointly with the emotion of a comment writer
\cite{Liu2013}.

\newcite{5286061} deal with writer's and reader's emotions on online
blogs and find that positive reader emotions tend to be linked to
positive writer emotions. \newcite{Buechel2017b} and
\newcite{buechel-hahn-2017-emobank} look into the effects of different
perspectives on annotation quality and find that the reader
perspective yields better inter-annotator agreement values.

\newcite{Haider2020} create an annotated corpus of poetry, in which
they make the task explicit that they care about the emotion perceived
by the reader, and not an emotion that is expressed by the author or a
character. They further propose that for the perception of art, the
commonly used set of fundamental emotions is not appropriate but
should be extended to a set of aesthetic emotions.

\section{Data Collection \& Annotation }

We gather the data in three steps: (1)~collecting the news and the reactions
they elicit in social media, (2)~filtering the resulting set to retain relevant
items, and (3)~sampling the final selection using various metrics.

The headlines are then annotated via crowdsourcing in two phases by
three annotators each in the first phase and by five annotators each
in the second phase. As a last step, the annotations are adjudicated
to form the gold standard. We describe each step in detail below.

\subsection{Collecting Headlines}

The first step consists of retrieving news headlines from the news publishers.
We further retrieve content related to a news item from social media: tweets
mentioning the headlines together with replies and Reddit posts that link to
the headlines. We use this additional information for subsampling described
later.

We manually select all news sources available as RSS feeds (82 out of
124) from the Media Bias Chart \cite{Otero2019}, a project that
analyzes reliability (from \textit{original fact reporting} to
\textit{containing inaccurate/fabricated information}) and political
bias (from \textit{most extreme left} to \textit{most extreme right})
of U.S. news sources. To have a source with a focus on more positive
emotions, we include Positive.News in addition.

Our news crawler retrieved daily headlines from the feeds, together with the
attached metadata (title, link, and summary of the news article) from March
2019 until October 2019. Every day, after the news collection finished, Twitter
was queried for 50 valid tweets for each headline\footnote{A tweet is
considered valid if it consists of more than 4 tokens which are not URLs,
hashtags, or user mentions.}. In addition to that, for each collected tweet, we
collect all valid replies and counts of being favorited, retweeted and replied
to in the first 24 hours after its publication.

The last step in the  pipeline is aquiring the top (``hot'') submissions in the
\texttt{/r/news}\footnote{\url{https://reddit.com/r/news}},
\texttt{/r/worldnews}\footnote{\url{https://reddit.com/r/worldnews}}
subreddits, and their metadata, including the number of up- and downvotes,
upvote ratio, number of comments, and the comments themselves.

\begin{table}[t]
\centering
\small
\renewcommand*{\arraystretch}{0.999}
\setlength\tabcolsep{1.3mm}
\begin{tabular}{lrrrrrr}
\toprule
Emotion& \rt{Random} & \rt{Entities} & \rt{NRC} & \rt{Reddit} & \rt{Twitter} & Total\\
\cmidrule(r){1-1}\cmidrule(l){2-2}\cmidrule(l){3-3}\cmidrule(l){4-4}\cmidrule(l){5-5}\cmidrule(l){6-6}\cmidrule(l){7-7}
Anger & 257 & 350 & 377 & 150 & 144 & 1278\\
Annoyance & 94 & 752 & 228 & 2 & 42 & 1118\\
Disgust & 125 & 98 & 89 & 31 & 50 & 392\\
Fear & 255 & 251 & 255 & 100 & 149 & 1010 \\
Guilt & 218 & 221 & 188 & 51 & 83 & 761\\
Joy & 122 & 104 & 95 & 70 & 68 & 459\\
Love & 6 & 51 & 20 & 0 & 4 & 81\\
Pessimism & 29 & 79 & 67 & 20 & 58 & 253 \\
Neg. Surprise & 351 & 352 & 412 & 216 & 367 & 1698\\
Optimism & 38 & 196 & 114 & 36 & 47 & 431\\
Pos. Surprise & 179 & 332 & 276 & 103 & 83 & 973\\
Pride & 17 & 111 & 42 & 12 & 17 & 199\\
Sadness & 186 & 251 & 281 & 203 & 158 & 1079\\
Shame & 112 & 154 & 140 & 44 & 114 & 564\\
Trust & 32 & 97 & 42 & 2 & 6 & 179\\
\cmidrule(r){1-1}\cmidrule(l){2-2}\cmidrule(l){3-3}\cmidrule(l){4-4}\cmidrule(l){5-5}\cmidrule(l){6-6}\cmidrule(l){7-7}
Total & 2021 & 3399 & 2626 & 1040 & 1390 & 10470 \\
\bottomrule
\end{tabular}
\caption{Sampling methods counts per adjudicated emotion.}
\label{tab:sampling}
\end{table}

\begin{table*}
\newbox\mybox
\setbox\mybox=\hbox{7. }
\centering
\small
\begin{tabularx}{\linewidth}{p{0.5mm}Xp{23mm}p{22mm}p{24mm}}
\toprule
& Question &  Type & Variable & Codes \\
\cmidrule(rl){2-2}\cmidrule(rl){3-3}\cmidrule(lr){4-4}\cmidrule(l){5-5}
\multirow{2}{*}{\rt{{\scriptsize Phase 1}}} 
& 1. Which emotion is most dominant in the given headline? & closed, single & Emotion & Emotions + None\\
& 2. Do you think the headline would stir up an emotion in readers? & closed, single & Emotion & Yes, No \\
\cmidrule(rl){2-2}\cmidrule(rl){3-3}\cmidrule(lr){4-4}\cmidrule(l){5-5}
\multirow{10}{*}{\rt{{\scriptsize Phase 2}}}
& 1. Which emotion is most dominant in the given headline? & closed, single & Emotion & Emotions \\
& 2. How intensely is the emotion expressed? &  closed, single & Intensity & Low, Med., High \\
& 3. Which words helped you in identifying the emotion? & open & Cue & String \\
& 4. Is the experiencer of the emotion mentioned? &  close & Experiencer & Yes, No \\
& 5. Who is the experiencer of the emotion? & open & Experiencer & String \\
& 6. Who or what is the emotion directed at? & open & Target & String \\
& 7. Select the words that explain what happened\par \hspace{\wd\mybox}that caused the expressed emotion. & open & Cause & String \\
& 8. Which other emotions are expressed in the given headline? & closed, multiple & Other Emotions  &  Emotions \\
& 9. Which emotion(s) did you feel while reading this headline? & closed, multiple & Reader Emotions  &  Emotions \\
\bottomrule
\end{tabularx}
\caption{Questionnaires for the two annotation phases. Emotions are Anger, Annoyance, Disgust, Fear, Guilt, Joy, Love, Pessimism, Neg. Surprise, Optimism, Negative Surprise, Optimism, Positive Surprise, Pride, Sadness, Shame, and Trust.}
\label{tab:questions}
\end{table*}

\subsection{Filtering \& Postprocessing}

We remove headlines that have less than 6 tokens (\eg, ``Small or
nothing'', ``But Her Emails'', ``Red for Higher Ed''), as well as those
starting with certain phrases, such as ``Ep.'',``Watch Live:'', ``Playlist:'',
``Guide to'', and ``Ten Things''. We also filter-out headlines that contain a
date (\eg, ``Headlines for March 15, 2019'') and words from the headlines which
refer to visual content (\eg ``video'', ``photo'', ``image'', ``graphic'',
``watch'').

\subsection{Sampling Headlines}
To aquire data across a wide political and stylistic spectrum, we
stratify the remaining headlines by source (150 headlines from each
source). We further subsample according to a set of different
strategies. From each strategy, we use the same number of
headlines. These are: 1)~randomly select headlines, 2)~select
headlines with high count of emotion terms, 3)~select headlines that
contain named entities, and 4)~select the headlines with high impact
on social media.

\paragraph{Random Sampling.} The goal of the first sampling method is to
collect a random sample of headlines that is representative and not biased
towards any source or content type. Note that the sample produced using this
strategy might not be as rich with emotional content as the other samples.

\paragraph{Sampling via NRC.} For the second sampling strategy, we hypothesize
that headlines containing emotionally charged words are also likely to contain
the structures we aim to annotate. This strategy selects headlines whose words
are in the NRC dictionary \cite{Mohammad2013}.

\paragraph{Sampling Entities.} We further hypothesize that headlines that
mention named entities may also contain experiencers or targets of emotions,
and therefore, they are likely to present a complete emotion structure.  This
sampling method yields headlines that contain at least one entity name,
according to the recognition from spaCy that is trained on OntoNotes 5 and
Wikipedia corpus.\footnote{\url{https://spacy.io/api/annotation}, last accessed
27 Nov 2019} We consider organization names, persons, nationalities, religious,
political groups, buildings, countries, and other locations.

\paragraph{Sampling based on Reddit \& Twitter.} The last sampling strategy
involves Twitter and Reddit metadata. This enables us to select and sample
headlines based on their impact on social media (under the assumption that this
correlates with the emotional connotation of the headline). This strategy chooses
them equally from the most favorited tweets, most retweeted headlines on
Twitter, most replied to tweets on Twitter, as well as most upvoted and most
commented on posts on Reddit.

\bigskip
Table~\ref{tab:sampling} on the previous page shows how many headlines are
selected by each sampling method in relation to the most dominant
emotion, which is the first of our annotation steps described in
Section~\ref{sec:Phase-1}

\begin{table*}
\centering
\small
\setlength\tabcolsep{1.8mm}
\begin{tabularx}{\linewidth}{lrrrrX}
\toprule
Rule &  Cue & Exp. & Cause &  Target  & Example \\
\cmidrule(r){1-1}\cmidrule(l){2-2}\cmidrule(l){3-3}\cmidrule(l){4-4}\cmidrule(l){5-5}\cmidrule(l){6-6}
 1. Majority & 3,872 & 4,820 & 3,678 & 3,308 & $\left(\textrm{span}_1 ; \textrm{span}_1 ; \textrm{span}_2\right) \rightarrow \textrm{span}{_1}$   \\
 2. Most common subsequence & 163 & 70 & 1,114 & 1,163  & \{$w_2$, $w_3$\}; \{$w_1$, $w_2$, $w_3$\}; \{$w_2$, $w_3$, $w_4$\}  $\rightarrow$ \{$w_2$, $w_3$\} \\
 3. Longest common subsequ. & 349 & 74 & 170 & 419 & \{$w_1$, $w_2$, $w_3$\}; \{$w_1$, $w_2$, $w_3$, $w_4$\}; \{$w_3$, $w_4$\} $\rightarrow$ \{$w_1$, $w_2$, $w_3$\} \\
 4. Noun Chunks   & 0 & 11 & 0 & 0 & \\
\cmidrule(r){1-1}\cmidrule(l){2-2}\cmidrule(l){3-3}\cmidrule(l){4-4}\cmidrule(l){5-5}\cmidrule(l){6-6}
 5. Manual & 611 & 25 & 38 & 110 &\\
\bottomrule
\end{tabularx}
\caption{Heuristics used in adjudicating gold corpus in the order of application on the questions of the type \textit{open} and their counts. $w_i$ refers to the the word with the index i in the headline, each set of words represents an annotation.}
\label{tab:heuristics-examples-numbers}
\end{table*}

\subsection{Annotation Procedure}

Using these sampling and filtering methods, we select 9,932 headlines. Next,
we set up two questionnaires (see Table~\ref{tab:questions}) for the two
annotation phases that we describe below. We use Figure
Eight\footnote{\url{https://figure-eight.com}, last accessed 27 Nov 2019}.

\subsubsection{Phase 1: Selecting Emotional Headlines}
\label{sec:Phase-1}

The first questionnaire is meant to determine the dominant emotion of a
headline if that exists, and whether the headline triggers an emotion in a
reader. We hypothesize that these two questions help us to retain only relevant
headlines for the next, more expensive, annotation phase.

During this phase, 9,932 headlines were annotated each by three
annotators. The first question of the first phase (P1Q1) is: ``Which
emotion is most dominant in the given headline?'' and annotators are
provided a closed list of 15 emotion categories to which the category
\emph{No emotion} was added. The second question (P1Q2) aims to answer
whether a given headline would stir up an emotion in most readers. The
annotators could choose one from only two possible answers (\emph{yes}
or \emph{no}, see Table~\ref{tab:questions} and
Figure~\ref{example-details} for details).

Our set of 15 emotion categories is an extended set over Plutchik's emotion
classes and comprises \textit{anger}, \textit{annoyance}, \textit{disgust},
\textit{fear}, \textit{guilt}, \textit{joy}, \textit{love}, \textit{pessimism},
\textit{negative surprise}, \textit{optimism}, \textit{positive surprise},
\textit{pride}, \textit{sadness}, \textit{shame}, and \textit{trust}. Such a
diverse set of emotion labels is meant to provide a more fine-grained analysis and
equip the annotators with a wider range of answer choices.

\subsubsection{Phase 2: Emotion and Role Annotation}

The annotations collected during the first phase are automatically ranked, and
the ranking is used to decide which headlines are further annotated in the
second phase. Ranking consists of sorting by agreement on P1Q1,
considering P1Q2 in the case of ties.

The top 5,000 ranked headlines are annotated by five annotators for
emotion class, intensity, reader emotion, and other emotions in case
there is not only one emotion. Along with these closed annotation
tasks, the annotators are asked to answer several open questions,
namely (1)~who is the experiencer of the emotion (if mentioned),
(2)~what event triggered the annotated emotion (if mentioned), (3)~if
the emotion had a target, and (4)~who or what is the target. The
annotators are free to select multiple instances related to the
dominant emotion by copy-paste into the answer field. For more details
on the exact questions and examples of answers, see
Table~\ref{tab:questions}. Figure~\ref{example-details} shows a
depiction of the procedure.

\subsubsection{Quality Control and Results}
To control the quality, we ensured that a single annotator annotates a maximum
of 120 headlines (this protects the annotators from reading too many news
headlines and from dominating the annotations). Secondly, we let only
annotators who geographically reside in the U.S. contribute to the task.

We test the annotators on a set of 1,100 test questions for the first
phase (about 10\% of the data) and 500 for the second
phase. Annotators were required to pass 95\%.
The questions were generated based on hand-picked non-ambiguous real headlines
through swapping out relevant words from the headline in order to obtain a
different annotation, for instance, for ``Djokovic happy to carry on cruising'',
we would swap ``Djokovic'' with a different entity, the cue ``happy'' to a
different emotion expression.

Further, we exclude Phase 1 annotations that were done in less than 10
seconds and Phase 2 annotations that were done in less than 70
seconds.

After we collected all annotations, we found unreliable annotators for both
phases in the following way: for each annotator and for each question, we
compute the probability with which the annotator agrees with the response
chosen by the majority. If the computed probability is more than two standard
deviations away from the mean, we discard all annotations done by that
annotator.

On average, 310 distinct annotators needed 15 seconds in the first phase. We
followed the guidelines of the platform regarding payment and decided to pay
for each judgment \$0.02 for Phase 1 (total of \$816.00). For the
second phase, 331 distinct annotators needed on average $\approx$1:17 minutes
to perform one judgment. Each judgment was paid with \$0.08 (total
\$2,720.00).

\subsection{Adjudication of Annotations}
In this section, we describe the adjudication process we undertook to create
the gold dataset and the difficulties we faced in creating a gold set out of
the collected annotations.

The first step was to discard wrong annotations for open questions, such as
annotations in languages other than English, or annotations of spans that were
not part of the headline. In the next step, we incrementally apply a set of
rules to the annotated instances in a one-or-nothing fashion. Specifically, we
incrementally test each instance for several criteria in such a way that if at
least one criterium is satisfied, the instance is accepted and its adjudication
is finalized. Instances that do not satisfy at least one criterium are
adjudicated manually by us.

\paragraph{Relative Majority Rule.} This filter is
applied to all questions regardless of their type. Effectively, whenever an
entire annotation is agreed upon by at least two annotators, we use all
parts of this annotation as the gold annotation.
Given the headline depicted in Figure~\ref{example-details} with the
following target role annotations by different annotators: \textit{``A
  couple'', ``None'', ``A couple'', ``officials'', ``their
  helicopter''}.  The resulting gold annotation is \textit{``A
  couple''} and the adjudication process for the target ends.

\paragraph{Most Common Subsequence Rule.} This rule is only applied to
open text questions. It takes the most common smallest string
intersection of all annotations. In the headline above, the
experiencer annotations \textit{``A couple'', ``infuriated
  officials'', ``officials'', ``officials'', ``infuriated officials''}
would lead to \textit{``officials''}.

\paragraph{Longest Common Subsequence Rule.} This rule is only applied if
two different intersections are the most common (previous rule), and these two
intersect. We then accept the longest common subsequence. Revisiting the example for deciding
on the \textit{cause} role with the annotations \textit{``by landing their
    helicopter in the nature reserve'', ``by landing their helicopter'',
``landing their helicopter in the nature reserve'', ``a couple infuriated
officials'', ``infuriated''} the adjudicated gold is \textit{``landing their
helicopter in the nature reserve''}.

Table~\ref{tab:heuristics-examples-numbers} shows through examples of how each
rule works and how many instances are ``solved'' by each adjudication rule.

\paragraph{Noun Chunks} For the role of experiencer, we accept only
the most-common noun-chunk(s)\footnote{We used spaCy's named entity recognition
model: \url{https://spacy.io/api/annotation\#named-entities}, last accessed Nov
25, 2019}.

\bigskip
The annotations that are left after being processed by all the rules
described above are being adjudicated manually by the authors of the
paper. We show examples for all roles in
Table~\ref{tab:entity-examples}.

\section{Analysis}

\subsection{Inter-Annotator Agreement}

We calculate the agreement on the full set of annotations from each phase for
the two question types, namely \emph{open} vs. \emph{closed}, where the first
deal with emotion classification and second with the roles \emph{cue},
\emph{experiencer}, \emph{cause}, and \emph{target}.

\subsubsection{Emotion}

\begin{table}
\centering
\small
\renewcommand*{\arraystretch}{0.9}
\setlength\tabcolsep{1.2mm}
\begin{tabularx}{\linewidth}{llX}
\toprule
Role & Chunk & Examples \\
\cmidrule(r){1-1}\cmidrule(lr){2-2}\cmidrule(l){3-3}
\multirow{2}{*}{Exp} & NP & cops, David Beckham, Florida National Park, Democrats, El Salvador's President, former Trump associate\\
& AdjP & illegal immigrant, muslim women from Sri Lanka, indian farmers, syrian woman, western media, dutch doctor\\
\cmidrule(r){1-1}\cmidrule(lr){2-2}\cmidrule(l){3-3}
\multirow{2}{*}{Cue} & NP & life lessons, scandal, no plans to stop, rebellion, record, sex assault \\ 
& AdjP & holy guacamole!, traumatized \\
& VP   & infuriates, fires, blasts, pushing, doing drugs, will shock \\
\cmidrule(r){1-1}\cmidrule(lr){2-2}\cmidrule(l){3-3}
\multirow{2}{*}{Cause} & VP & escaping the dictatorship of the dollar, giving birth in the wake of a storm \\
& Clause & pensioners being forced to sell their home to pay for care \\
& NP & trump tax law, trade war, theory of change at first democratic debate, two armed men \\
\cmidrule(r){1-1}\cmidrule(lr){2-2}\cmidrule(l){3-3}
\multirow{2}{*}{Target} & AdvP & lazy students\\
& NP & nebraska flood victims, immigrant detention centers, measles crisis \\
\bottomrule
\end{tabularx}%
\caption{Example of linguistic realizations of the different roles.}
\label{tab:entity-examples}
\end{table}

\begin{table}
\centering\small
\begin{tabular}{lcccccc}
\toprule
{Agreement} & \rt{Emo./Non-Emo.} & \rt{Reader Percep.} & \rt{Dominant Emo.} & \rt{Intensity} & \rt{Other Emotions} & \rt{Reader Emotions} \\
\cmidrule(r){1-1}\cmidrule(rl){2-2}\cmidrule(rl){3-3}\cmidrule(rl){4-4}\cmidrule(rl){5-5}\cmidrule(rl){6-6}\cmidrule(l){7-7}
$\kappa$ & 0.34 & 0.09 & 0.09 & 0.22 & 0.06 & 0.05   \\
\% & 0.71 & 0.69 & 0.17 & 0.92 & 0.80 & 0.80  \\
H (in bits) & 0.40 & 0.42 & 1.74 & 0.13 & 0.36 & 0.37  \\
\bottomrule
\end{tabular}
\caption{Agreement statistics on closed questions. Comparing with the questions in Table~\ref{tab:questions}, Emotional/Non-Emotional uses the annotations of Phase 1 Question 1 (P1Q1). In the same way, Reader perception refers to P1Q2, Dominant Emotion is P2Q1, Intensity is linked to P2Q2, Other Emotions to P2Q8, and Reader Emotions to P2Q9.}
\label{tab:emotion-agreement}
\end{table}

\begin{table}
  \centering
  \small
  \begin{tabular}{lllll}
    \toprule
    & \multicolumn{4}{c}{\# of annotators agreeing}\\ 
    \cmidrule{2-5}
    Emotion  & $\geq$ 2 & $\geq$ 3 & $\geq$ 4 &  $\geq$ 5 \\
    \cmidrule(r){1-1}\cmidrule(rl){2-2}\cmidrule(rl){3-3}\cmidrule(rl){4-4}\cmidrule(l){5-5}
    Anger & 1.00 & 0.74 & 0.33 & 0.15\\
    Annoyance & 1.00 & 0.71 & 0.22 & 0.05\\
    Disgust & 1.00 & 0.78 & 0.21 & 0.08\\
    Fear & 1.00 & 0.83 & 0.44 & 0.23\\
    Guilt & 1.00 & 0.82 & 0.37 & 0.14\\
    Joy & 1.00 & 0.84 & 0.43 & 0.17\\
    Love & 1.00 & 0.90 & 0.62 & 0.48\\
    Pessimism & 1.00 & 0.76 & 0.24 & 0.07\\
    Neg. Surprise & 1.00 & 0.81 & 0.32 & 0.11\\
    Optimism & 1.00 & 0.69 & 0.31 & 0.12\\
    Pos. Surprise & 1.00 & 0.82 & 0.38 & 0.14\\
    Pride & 1.00 & 0.70 & 0.30 & 0.26\\
    Sadness & 1.00 & 0.86 & 0.50 & 0.24\\
    Shame & 1.00 & 0.63 & 0.24 & 0.13\\
    Trust & 1.00 & 0.43 & 0.05 & 0.05\\
    \cmidrule(r){1-1}\cmidrule(rl){2-2}\cmidrule(rl){3-3}\cmidrule(rl){4-4}\cmidrule(l){5-5}
    Micro Average & 1.00 & 0.75 & 0.33 & 0.16\\
    \bottomrule
  \end{tabular}
  \caption{Percentage agreement per emotion category on most dominant
    emotion (second phase). Each column shows the percentage of emotions for
    which the \# of annotators agreeing is greater than 2, 3, 4, and
    5}
  \label{tab:majority}
\end{table}

\begin{table}
\centering\small
  \begin{tabular}{lccccccc}
  \toprule
    Type & $\kappa$ & \F &  \% & Tok. & MASI & H\\
    \cmidrule(r){1-1}\cmidrule(lr){2-2}\cmidrule(lr){3-3}\cmidrule(lr){4-4}\cmidrule(lr){5-5}\cmidrule(lr){6-6}\cmidrule(l){7-7}
    Experiencer & .40 & .43 & .36 & .56 & .35 & .72\\
    Cue & .31 & .39 & .30 & .73 & .55 & .94 \\
    Cause & .28 & .60 & .16 & .58 & .47 & .58\\
    Target & .15 & .36 & .12 & .45 & .54 & .04 \\
    \bottomrule
  \end{tabular}
  \caption{Pairwise inter-annotator agreement (mean) for the open questions annotations. We report for each role the following scores: Fleiss's $\kappa$,  Accuracy, \F score, Proportional Token Overlap, MASI and Entropy}
  \label{tab:roles-agreement}
\end{table}

\begin{table*}
\small
\newcommand{\sep}{\cmidrule(r){1-1}\cmidrule(r){2-2}\cmidrule(r){3-3}\cmidrule(r){4-4}\cmidrule(r){5-5}\cmidrule(r){6-6}\cmidrule(r){7-7}\cmidrule(r){8-8}\cmidrule(r){9-9}\cmidrule(r){10-10}\cmidrule(r){11-11}\cmidrule(r){12-12}\cmidrule(r){13-13}\cmidrule(r){14-14}\cmidrule(r){15-15}\cmidrule(r){16-16}\cmidrule(r){17-17}\cmidrule(r){18-18}\cmidrule(r){19-19}}
\renewcommand*{\arraystretch}{1}
\setlength\tabcolsep{1.2mm}
\centering
\begin{tabular}{lrrrrrrrrrrrrrrrr|rr}
\toprule
\multicolumn{17}{c}{Dominant Emotion} &  \multicolumn{2}{c}{Anno.} \\
\cmidrule(lr){2-17}\cmidrule(rl){18-19}
Role &  \rt{Anger} & \rt{Annoyance} &
\rt{Disgust} & \rt{Fear} & \rt{Guilt} & \rt{Joy} & \rt{Love} &
\rt{Pessimism} & \rt{Neg. Surprise}  &\rt{Optimism} & \rt{Pos. Surprise} & \rt{Pride} & \rt{Sadness} & \rt{Shame} & \rt{Trust} & Total & \rt{Mean Tok. } & \rt{Std. Dev Tok.}\\
\sep
Experiencer & 371 & 214 & 292 & 294 & 144 & 176 & 39 & 231 & 628 & 212 & 391 & 52 & 238 & 89 & 95 & 3466 & 1.96 & 1.00\\
Cue & 454 & 342 & 371 & 410 & 175 & 256 & 62 & 315 & 873 & 307 & 569 & 60 & 383 & 117 & 120 & 4814 & 1.45 & 1.10\\
Cause & 449 & 341 & 375 & 408 & 171 & 260 & 58 & 315 & 871 & 310 & 562 & 65 & 376 & 118 & 119 & 4798 & 7.21 & 3.81\\
Target & 428 & 319 & 356 & 383 & 164 & 227 & 54 & 297 & 805 & 289 & 529 & 60 & 338 & 111 & 117 & 4477 & 4.67 & 3.56\\
\sep
Overall & 1702 & 1216 & 1394 & 1495 & 654 & 919 & 213 & 1158 & 3177 & 1118 & 2051 & 237 & 1335 & 435 & 451 & 17555 & 3.94 & 3.64\\
\bottomrule
\end{tabular}%
\caption{Corpus statistics for role annotations. Columns indicate
  how frequent the respective emotions are in relation to the annotated role and annotation length.}
\label{tab:general}
\end{table*}

We use Fleiss' Kappa ($\kappa$) to measure the inter-annotator agreement for
closed questions \cite{Poesio2008,Fleiss2013}. Besides, we report the average
percentage of overlaps between all pairs of annotators (\%) and the mean
entropy of annotations in bits. Higher agreement correlates with lower entropy.
As Table~\ref{tab:emotion-agreement} shows, the agreement on the question
whether a headline is emotional or not obtains the highest agreement (.34),
followed by the question on intensity (.22). The lowest agreement is on the
question to find the most dominant emotion (.09).

All metrics show comparably low agreement on the closed questions, especially
on the question of the most dominant emotion. This is reasonable, given that
emotion annotation is an ambiguous, subjective, and difficult task. This aspect
lead to the decision of not purely calculating a majority vote label but to
consider the diversity in human interpretation of emotion categories and
publish the annotations by all annotators.

Table~\ref{tab:majority} shows the counts of annotators agreeing on a
particular emotion. We observe that \emph{Love}, \emph{Pride}, and
\emph{Sadness} show highest intersubjectivity followed closely by \emph{Fear}
and \emph{Joy}. \emph{Anger} and \emph{Annoyance} show, given their similarity,
lower scores. Note that the micro average of the basic emotions (+ love) is
.21 for when more than five annotators agree.

\subsubsection{Roles}

Table~\ref{tab:roles-agreement} presents the mean of pair-wise inter-annotator
agreement for each role. We report average pair-wise Fleiss' $\kappa$,
span-based exact \F over the annotated spans, accuracy, proportional token
overlap, and the measure of agreement on set-valued items, MASI
\cite{Passonneau2004}.
We observe a fair agreement on the open annotation tasks. The highest agreement
is for the role of the \emph{Experiencer}, followed by \emph{Cue},
\emph{Cause}, and \emph{Target}.

This seems to correlate with the length of the annotated spans (see
Table~\ref{tab:general}). This finding is consistent with \newcite{Kim2018}.
Presumably, \emph{Experiencers} are easier to annotate as they often are noun
phrases whereas causes can be convoluted relative clauses.

\subsection{General Corpus Statistics}
In the following, we report numbers of the adjudicated data set for
simplicity of discussion. Please note that we publish all annotations
by all annotators and suggest that computational models should
consider the distribution of annotations instead of one adjudicated
gold. The latter would be a simplification which we consider to not be
appropriate.

\textit{Good\-News\-Every\-one} contains 5,000 headlines from various
news sources. Overall, the corpus is composed of 56,612 words (354,173
characters), out of which 17,513 are unique. The headline length is
short, with 11 words on average. The shortest headline contains six
words, while the longest headline contains 32 words.  The length of a
headline in characters ranges from 24 the shortest to 199 the longest.

Table~\ref{tab:general} presents the total number of adjudicated
annotations for each role in relation to the dominant
emotion. \textit{Good\-News\-Everyone} consists of 5,000 headlines,
3,312 of which have an annotated dominant emotion via majority
vote. The rest of the 1,688 headlines (up to 5,000) ended in ties for
the most dominant emotion category and were adjudicated manually. The
emotion category \textit{Negative Surprise} has the highest number of
annotations, while \textit{Love} has the lowest number of
annotations. In most cases, \textit{Cues} are single tokens (\eg,
``infuriates'', ``slams''), \textit{Causes} have the largest
proportion of annotations that span more than seven tokens on average
(65\% out of all annotations in this category).

\begin{figure}
 \centering
 \includegraphics[scale=1.05]{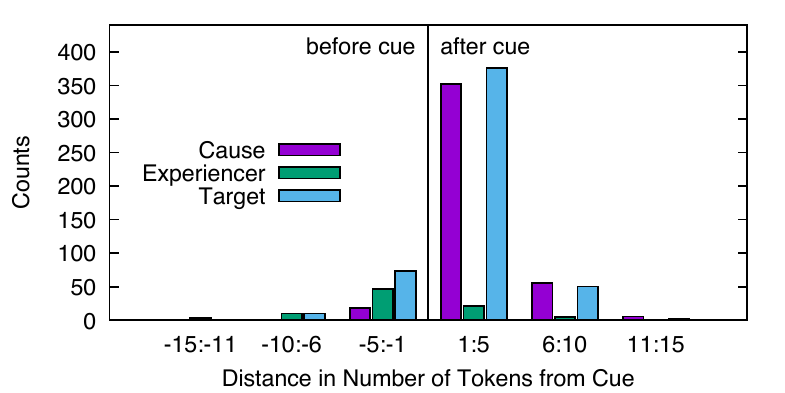}
 \caption{Distances between emotion cues and the other relations: cause, experiencer, and target.}
 \label{fig:distogram}
\end{figure}

For the role of \textit{Experiencer}, we see the lowest number of
annotations (19\%), which is a very different result to the one
presented by \newcite{Kim2018}, where the role \emph{Experiencer} was
the most annotated. We hypothesize that this is the effect of the
domain we annotated; it is more likely to encounter explicit
experiencers in literature (as literary characters) than in news
headlines. As we can see, the \textit{Cue} and the \textit{Cause}
relations dominate the dataset (27\% each), followed by
\textit{Target} (25\%) relations.

Table \ref{tab:general} also shows how many times each emotion
triggered a certain relation. In this sense, \textit{Negative
  Surprise} and \textit{Positive Surprise} has triggered the most
\textit{Experiencer}, and \textit{Cause} and \textit{Target}
relations, which due to the prevalence of the annotations for this
emotion in the dataset.

Further, Figure~\ref{fig:distogram}, shows the distances of the
different roles from the cue. The causes and targets are predominantly
realized right of the cue, while the experiencer occurs more often
left of the cue.

\subsection{Emotions across News Sources}

\begin{table*}
\renewcommand*{\arraystretch}{1.0}
\newcommand{\signstar}{\textsuperscript{\faAsterisk}}
\setlength\tabcolsep{1mm}
\centering
\footnotesize
\begin{tabular}{lll}
\toprule
Emotion & Dominant Emotion & Reader Emotions \\
\cmidrule(r){1-1}\cmidrule(rl){2-2}\cmidrule(l){3-3}
Anger & The Blaze, The Daily Wire, BuzzFeed & The Gateway Pundit, The
Daily Mail, Talking Points Memo \\
Annoyance & Vice, NewsBusters, AlterNet & Vice, The Week, Business Insider \\
Disgust & BuzzFeed, The Hill, NewsBusters & Mother Jones, The Blaze,
Daily Caller \\
Fear & The Daily Mail, Los Angeles Times, BBC & Palmer Report, CNN, InfoWars \\
Guilt & Fox News, The Daily Mail, Vice & The Washington Times, Reason,
National Review \\
Joy & Time, Positive.News, BBC & Positive.News, ThinkProgress, AlterNet \\
Love & Positive.News, The New Yorker, BBC & Positive.News, AlterNet, Twitchy \\
Pessimism & MotherJones, Intercept, Financial Times & The Guardian,
Truthout, The Washinghton Post \\
Neg. Surprise & The Daily Mail, MarketWatch, Vice & The Daily Mail, BBC, Breitbart \\

Optimism & Bussines Insider, The Week, The Fiscal Times & MarketWatch,
Positive.News, The New Republic \\
Pos. Surprise & Positive.News, BBC, MarketWatch & Positive.News, The
Washington Post, MotherJones \\
Pride & Positive.News, The Guardian, The New Yorker & Daily Kos, NBC,
The Guardian \\
Sadness & The Daily Mail, CNN, Daily Caller & The Daily Mail, CNN, The
Washington Post \\
Shame & The Daily Mail, The Guardian, The Daily Wire & Mother Jones,
National Review, Fox News \\
Trust & The Daily Signal, Fox News, Mother Jones & Economist, The Los
Angeles Times, The Hill \\
\bottomrule
\end{tabular}
\caption{Top three media sources in relation to the main emotion in
  the text and the reader's emotion.}
\label{tab:media}
\end{table*}

Table~\ref{tab:media} shows the top three media sources for each
emotion that has been annotated to be the dominating one and the
respective sources for the reader's emotion.

Unsurprisingly for the positive emotions, \emph{Joy}, \emph{Love},
\emph{Positive Surprise}, and \emph{Pride} there is one common source,
namely Positive.News. For strong negative emotions such as
\emph{Anger} and \emph{Disgust} the top three across the different
emotions vary.

Though the annotated data for each of the sources is comparably
limited, there are a set of interesting findings. Infowars, which the
Media Bias Chart categorizes as most right wing and least reliable is
found in the list of most frequently being associated with \emph{Fear}
in the reader. Breitbart is found to be associated with \emph{Negative
  Surprise} in the reader. However, both these sources are not in the
list of the text-level emotion annotation. Surprisingly, BBC and LA
Times are in the list of the most associated with fear on the
text-level, despite of both sources being relatively neutral and
moderately factual.  Further, it is noteworthy that Reuters, ABC News,
as being categorized as maximally reliable, are not in the top emotion
list at all.

This analysis regarding emotions and media sources is also interesting
the other way round, namely to check which emotions are dominating
which source. From all sources we have in our corpus, nearly all of
them have their headlines predominantly annotated with surprise,
either negative or positive. That could be expected, given that news
headlines often communicate something that has not been
known. Exceptions are \emph{Buzzfeed} and \emph{The Hill}, which are
dominated by disgust, \emph{CNN}, \emph{Fox News}, \emph{Washington
  Post}, \emph{The Advocate}, all dominated by \emph{Sadness}, and
\emph{Economist}, \emph{Financial Times}, \emph{MotherJones}, all
dominated either by \emph{Positive} or \emph{Negative
  Anticipation}. Only \emph{Time} has most headlines annotated as
\emph{Joy}.

Note that this analysis does not say a lot about what the media
sources publish -- it might also reflect on our sampling strategy and
point out what is discussed in social media or which headlines contain
emotion words from a dictionary.

\section{Baseline}

As an estimate for the difficulty of the task, we provide baseline
results. We focus on the segmentation tasks as these form the main
novel contribution of our data set. Therefore, we formulate the task
as sequence labeling of emotion cues, mentions of experiencers,
targets, and causes with a bidirectional long short-term memory
networks with a CRF layer (biLSTM-CRF) that uses ELMo embeddings
\cite{peters-etal-2018-deep} as input and an IOB alphabet as
output.

The results are shown in Table~\ref{tab:results}. We observe that the
results for the detection of experiencers performs best, with .48\F,
followed by the detection of causes with .37\F. The recognition of
causes and targets is more challenging, with .14\F and .09\F. Given
that these elements consist of longer spans, this is not too
surprising. These results are in line with the findings by
\newcite{Kim2018}, who report an acceptable result of .3\F for
experiencers and a low .06\F for targets. They were not able achieve
any correct segmentation prediction for causes, in contrast to our
experiment.

\begin{table}
\centering
\renewcommand*{\arraystretch}{1.0}
\newcommand{\signstar}{\textsuperscript{\faAsterisk}}
\setlength\tabcolsep{1.2mm}
\begin{tabular}{lccc}
\toprule
Category & P & R & \F \\
\cmidrule(r){1-1}\cmidrule(rl){2-2}\cmidrule(lr){3-3}\cmidrule(l){4-4}
Experiencer &0.44 & 0.53 & 0.48 \\
Cue   & 0.39 & 0.35 &  0.37\\
Cause & 0.19 & 0.11 & 0.14  \\
Target & 0.10 & 0.08 & 0.09 \\
\bottomrule
\end{tabular}%
\caption{Results for the baseline experiments.}
\label{tab:results}
\end{table}

\section{Conclusion and Future Work}
We introduce \textit{GoodNewsEveryone}, a
corpus of 5,000 headlines annotated for emotion categories, semantic roles,
and reader perspective. Such a dataset enables answering instance-based
questions, such as, ``who is experiencing what emotion and why?'' or more
general questions, like ``what are typical causes of joy in media?''.
To annotate the headlines, we employ a two-phase procedure and use
crowdsourcing. To obtain a gold dataset, we aggregate the annotations through
automatic heuristics.

As the evaluation of the inter-annotator agreement and the baseline
model results show, the task of annotating structures encompassing
emotions with the corresponding roles is a difficult one. We also note
that developing such a resource via crowdsourcing has its limitations,
due to the subjective nature of emotions, it is very challenging to
come up with an annotation methodology that would ensure less
dissenting annotations for the domain of headlines.

We release the raw dataset including all annotations by all
annotators, the aggregated gold dataset, and the questionnaires. The
released dataset will be useful for social science scholars, since it
contains valuable information about the interactions of emotions in
news headlines, and gives exciting insights into the language of
emotion expression in media. Finally, we would like to note that this
dataset is also useful to test structured prediction models in
general.

\section{Acknowledgements}

This research has been conducted within the CRETA project
(\url{http://www.creta.uni-stuttgart.de/}) which is funded by the
German Ministry for Education and Research (BMBF) and partially funded
by the German Research Council (DFG), projects SEAT (Structured
Multi-Domain Emotion Analysis from Text, KL 2869/1-1).  We thank
Enrica Troiano and Jeremy Barnes for fruitful discussions.

\def\UrlBreaks{\do\/\do-}

\urlstyle{same}

\section{Bibliographical References}

\end{document}